\documentclass[runningheads]{llncs}

\usepackage{eccv}

\usepackage{eccvabbrv}
\usepackage{caption}
\usepackage{graphicx}
\usepackage{booktabs}
\usepackage{makecell}
\usepackage[accsupp]{axessibility}  
\usepackage{multirow}
\usepackage{xcolor}
\usepackage{booktabs}
\usepackage{colortbl}
\usepackage{pifont}
\newcommand{\cmark}{\ding{51}}
\newcommand{\xmark}{\ding{55}}

\usepackage{hyperref}

\usepackage{orcidlink}

\begin{document}

\title{UHD-MFF: Shattering Barriers in Multi-Focus Ultra-High-Definition Image Fusion via Learnable Lookup Tables}

\titlerunning{UHD-MFF}

\author{Yibing Zhang\inst{1}\orcidlink{0009-0004-5234-2093}\thanks{Equal contribution.} \and
Xunpeng Yi\inst{1}\orcidlink{0000-0003-0116-3234}\protect\footnotemark[1] \and
Qinglong Yan\inst{1}\orcidlink{0009-0000-9553-8628} \and
Yeda Wang\inst{1}\orcidlink{0009-0005-6145-510X} \and
Han Xu\inst{3}\orcidlink{0000-0002-6291-2924} \and
Jiayi Ma\inst{1,2}\orcidlink{0000-0003-3264-3265}\thanks{Corresponding author.}}

\authorrunning{Y. Zhang et al.}

\institute{Electronic Information School, Wuhan University, Wuhan 430072, China \and
School of Robotics, Wuhan University, Wuhan 430072, China \and
School of Automation, Southeast University, Nanjing 210096, China
\email{\{zhangyibing,yixunpeng,qinglong\_yan,wangyeda\}@whu.edu.cn}\\
\email{xu\_han@seu.edu.cn, jyma2010@gmail.com}}

\maketitle

\begin{abstract}
  With the advancement of imaging technology, ultra-high-definition images have become increasingly essential in modern visual applications. However, existing multi-focus image fusion remains largely confined to low-resolution images and faces three major barriers in UHD scenarios, namely data availability, model adaptability, and deployment feasibility, which severely hinder its practical application. To shatter these barriers, first, we propose the UHD-MFF dataset, the first large-scale ultra-high-resolution multi-focus fusion dataset. Second, we propose a scale-specialized lookup-table framework tailored for ultra-high-resolution images, termed as UMF-LUT. It consists of Coarse-Region Lookup Table (C-LUT) and Detail-Edge Lookup Table (D-LUT). Specifically, C-LUT performs joint queries of multiple gradient cues and semantic cues at low-resolution scales to enable region-level decision-making. Also, D-LUT operates at high-resolution scales, leveraging efficient Laplacian cues to provide complementary edge-level decision information. Such a design makes the model particularly well-suited for ultra-high-resolution multi-focus image fusion. Finally, it offers strong deployability with minimal computational overhead, enabling real-time 4K multi-focus fusion and showing promising potential for smartphone. Extensive experiments demonstrate that it outperforms SOTA methods in both visual fidelity and quantitative metrics. It effectively advances the development of multi-focus image fusion toward ultra-high-resolution imaging scenarios. The code is available at \url{https://github.com/zyb5/UHD-MFF}.
\end{abstract}

\section{Introduction}
\label{sec:intro}

Due to the limited depth of field of optical lenses, a single image typically captures only part of a scene in focus, leaving other regions blurred and resulting in information loss. In this background, Multi-Focus Image Fusion (MFIF) has been developed as an effective solution~\cite{li2017multi-lowrank}. As an important branch of image fusion, MFIF combines multiple images captured at different focal lengths to generate a single all-in-focus image, thereby enhancing scene clarity and preserving more detail~\cite{liu2021learning,huang2022reconet}. Owing to its ability to handle complex scenes with varying focal distances, MFIF has been widely applied in areas such as macro digital photography.

\begin{figure*}[t]
\centering
\includegraphics[width=1\linewidth]{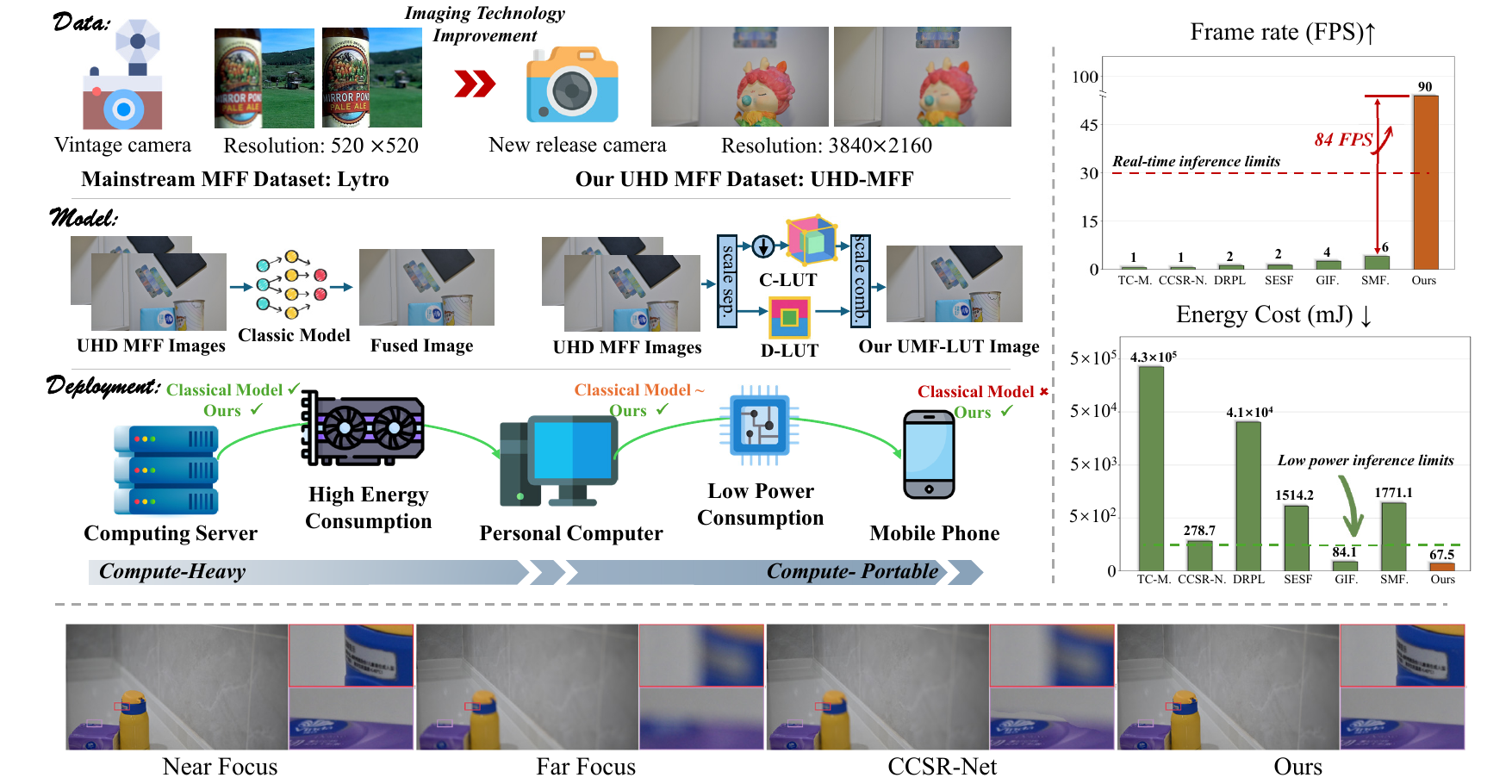}
\vspace{-0.2in}
\caption{Three primary barriers in ultra-high-resolution multi-focus image fusion. Our method is compared to state-of-the-art methods in terms of inference frame rate, energy cost, and visual quality.}
\label{fig:1-1}
\end{figure*}

MFIF has achieved substantial development in recent years~\cite{bhat2021multi-techsurvey,hu2023zmff,ouyang2025fusiongcn}. However, with the rapid advancement of imaging devices, ultra-high-resolution images have become the prevailing trend. Under this circumstance, the development of multi-focus image fusion faces new bottlenecks. As shown in Fig.~\ref{fig:1-1}, we summarize these challenges into three major barriers: \textbf{1) Data availability:} Existing mainstream datasets, such as Lytro and Real-MFF, are all limited to resolutions below 1080P. Such relatively low-resolution data are insufficient to support research on ultra-high-resolution image fusion. Moreover, models trained on low-resolution datasets often exhibit suboptimal performance when generalized to ultra-high-resolution scenarios. \textbf{2) Model adaptability:} Due to the lack of explicit consideration of ultra-high-resolution image scales, previous methods have often relied on conventional convolutional operations or attention mechanisms in a straightforward manner, without specialized adaptation for UHD images. \textbf{3) Deployment feasibility:} The demand for multi-focus image fusion on mobile platforms is substantial. However, existing methods generally do not support deployment on smartphones and, in some cases, are not even feasible for personal computers~\cite{yi2025lut, yi2024diff}. Instead, they rely on computation-intensive servers for execution, which severely restricts their practical applicability. These three barriers collectively impose significant constraints on the advancement of MFIF, hindering its ability to meet the demands of next-generation ultra-high-resolution imaging.

To address these issues, in terms of data availability, to support the advancement of ultra-high-definition multi-focus image fusion and fill the existing data gap in this field, we propose the UHD-MFF dataset, which comprises 1,950 pairs of near- and far-focus images at the resolution of $3840\times 2160$, encompassing a diverse range of indoor and outdoor scenes with varying degrees of focus. In terms of model adaptability, we propose the scale-specialized lookup-table multi-focus image fusion for ultra-high-resolution images, termed as UMF-LUT. It customizes the large-scale challenge of ultra-high-resolution multi-focus images by performing region-wise decisions at low-resolution scales and edge refinement at high-resolution scales. Firstly, at the low-resolution scales, we utilize multiple gradient cues, including gradient difference cues, dilated gradient cues, maximum gradient cues, and learnable regional semantic cues, as lookup elements in the coarse-region lookup table (C-LUT) to perform the region-wise decisions. Later, as a complement, at the high-resolution scales, we employ the Laplacian operator as a residual decision cue in the lightweight detail-edge lookup table (D-LUT) to achieve edge-level decision-making. This design not only ensures clear focus boundaries but also cleverly avoids the substantial memory overhead associated with full-resolution convolutions. It achieves strong multi-focus fusion performance through a scale-specific, customized combination of operations, making it particularly well-suited for high-resolution images. From the perspective of deployment feasibility, UMF-LUT incurs very low computational overhead, enabling real-time fusion of 4K-resolution UHD images on commercial GPUs and demonstrating potential for deployment on mobile smartphones.

Overall, our contributions can be summarized as follows:
\begin{itemize}
    \item \textbf{Data availability:} We propose UHD-MFF, the first large-scale ultra-high-resolution multi-focus image fusion dataset comprising 1,950 diverse image pairs. This fills the critical gap in UHD benchmarks and provides a robust foundation for future data-driven research.

\vspace{2mm}
    \item \textbf{Model adaptability:} We introduce UMF-LUT, a scale-decoupled architecture tailored for UHD images. By combining low-resolution regional decisions with native-resolution edge refinement, it ensures accurate focus perception while bypassing the memory bottlenecks of full-resolution convolutions.

\vspace{2mm}
    \item \textbf{Deployment feasibility:} Empowered by its special LUT design, UMF-LUT achieves real-time UHD inference with minimal computational overhead. Furthermore, we successfully demonstrate its deployment on mobile smartphones, breaking the hardware barrier of existing methods.

\end{itemize}

\section{Related Work}

\textbf{Traditional Methods.} Generally, traditional multi-focus image fusion algorithms can be categorized into two streams~\cite{liu2020multi-survey,bhat2021multi-techsurvey}: spatial domain-based methods and transform domain-based methods. Spatial domain methods directly manipulate pixels within source images to generate weight maps, which are then employed to fuse the pixels. In contrast, transform domain methods first transform source images into the frequency domain using specific decomposition rules. The decomposed coefficients are fused based on certain strategies, followed by an inverse transformation to reconstruct the final image. The quadtree-based method~\cite{bai2015quadtree} utilizes quadtree decomposition and weighted focus measures, which adaptively partitions the image into blocks of varying sizes to precisely extract focused regions. DSIFT~\cite{liu2015multi} innovatively introduces dense SIFT descriptors into the MFIF domain, employing them as activity measures for image patches to generate initial decision maps, which are further refined through feature matching. GFDF~\cite{qiu2019guided} leverages focus region detection and the edge-preserving properties of guided filtering to optimize coarse initial decision maps and correct boundary artifacts. Furthermore, the HOSVD and edge-intensity method~\cite{luo2017multi} models higher-order patch correlations. The JSR-based method~\cite{ma2019multi-sparse} introduces a fusion framework based on Joint Sparse Representation and optimization theory, decomposing source images into a common component containing redundant information and a component containing focused details.

\vspace{2mm}
\noindent
\textbf{Deep Learning-based Methods.} Deep learning has significantly advanced MFIF~\cite{zhang2021image-survey,wang2025mmae,yi2025artificial}, with existing methods primarily categorized into regression-based~\cite{zhang2020ifcnn} and classification-based approaches~\cite{shao2024stcu,quan2025dmanet}.
Addressing the lack of ground truth, SMFuse~\cite{ma2021smfuse} utilizes focus measures as auxiliary signals for self-supervised pixel-level focus detection via iterative optimization. MFF-GAN~\cite{zhang2021mff} employs unsupervised adversarial training with adaptive gradient constraints to capture textures while preserving source intensities. To enhance feature representation, MSFIN~\cite{liu2021multiscale} introduces a multi-scale interaction network to facilitate bidirectional complementarity between deep semantics and shallow details. For generalization, ZMFF~\cite{hu2023zmff} adopts a zero-shot framework optimizing loss functions directly from inputs to eliminate train-test distribution shifts. Bridging model- and learning-based methods, CCSR-Net~\cite{zheng2025unfolding} unrolls the iterative solution of coupled convolutional sparse representation into a neural network, combining theoretical interpretability with efficient inference. Several unified frameworks also incorporate the MFIF task. U2Fusion~\cite{xu2020u2fusion} leverages adaptive information preservation metrics to automatically estimate pixel importance and retain critical structures. More recently, TC-MoA~\cite{Zhu_2024_CVPR} introduces a task-customized Mixture of Adapters for general image fusion.

\section{Our Approach}

\subsection{Scale-Specialized Lookup-Table}
\label{subsec2}

\begin{figure*}[t]
\centering
\includegraphics[width=1\textwidth]{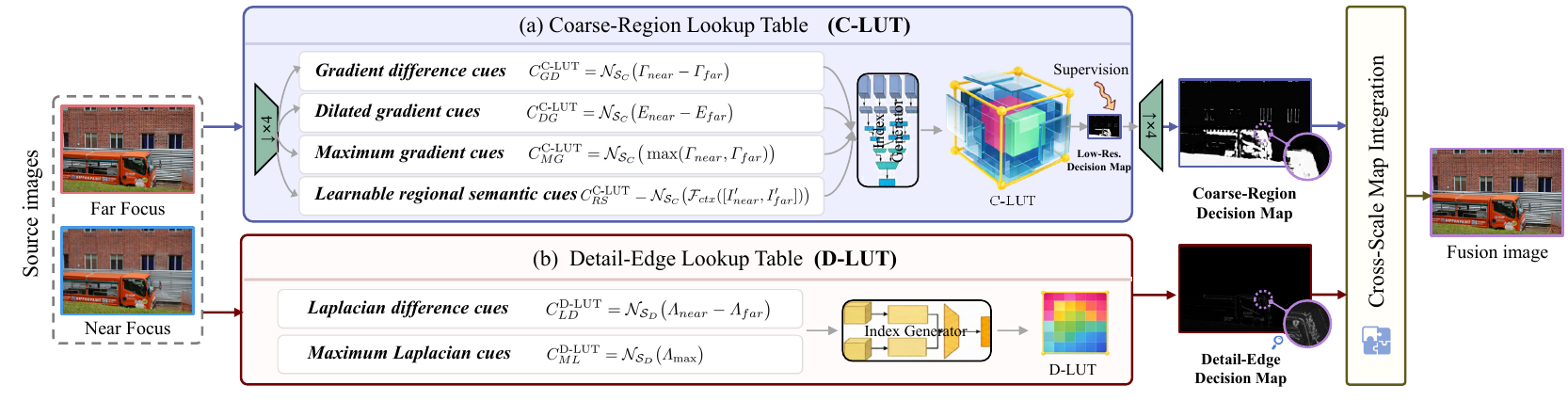}
\caption{Overall Structure of the proposed UMF-LUT.}
\label{fig:3-1}
\end{figure*}

\noindent
\textbf{Coarse-Region Lookup Table (C-LUT):} To expand the receptive field while mitigating computational overhead across UHD scales, spatial compression is applied to the Coarse-Region Lookup Table in UMF-LUT, as shown in Fig.~\ref{fig:3-1}. Given the source image pair $I_{near}, I_{far} \in \mathbb{R}^{H \times W \times C}$, a down-sampling operator $\mathcal{D}_{\downarrow s}(\cdot)$ with a scale factor $s$ projects the images into a low-resolution space:
\begin{equation}
\label{eq:downsample}
    I'_{k} = \mathcal{D}_{\downarrow s}(I_{k}), \quad k \in \{near, far\}.
\end{equation}

The primary objective of this coarse-scale operation is to establish a reliable regional consensus regarding focus attributes. To ensure computational efficiency, this complex decision-making process is mapped into a lookup table constructed from informative cues. Because focus levels are intrinsically correlated with regional texture variations, first-order Sobel operators are employed in the C-LUT to extract base gradient cues:
\begin{equation}
    \Gamma_k = |\mathcal{H}_x \ast I_k'| + |\mathcal{H}_y \ast I_k'|, \quad k \in \{near, far\},
\end{equation}
where $\mathcal{H}_x$ and $\mathcal{H}_y$ represent the horizontal and vertical Sobel operator kernels, and $\ast$ denotes the convolution operation.

To strictly map these continuous cues into the discrete index space of the C-LUT while maximizing the utilization of the lookup capacity, we introduce a unified spatial normalization operator $\mathcal{N}_{\mathcal{S}_C}(\cdot)$. This operator applies a non-linear activation followed by a scale-and-shift transformation, bounding any continuous input within the grid range $[0, \mathcal{S}_C-1]$. The resulting index group $\mathbf{c} = \{c_1, c_2, c_3, c_4\}$ is formulated as follows:

\vspace{2mm}
\noindent
\textbf{\textit{1) Gradient difference cues:}} This cue measures the local sharpness difference between the two images and directly reflects which image exhibits stronger structural responses:
\begin{equation}
    C^{\text{C-LUT}}_{GD} = \mathcal{N}_{\mathcal{S}_C}\big(\Gamma_{near} - \Gamma_{far}\big).
\end{equation}

\noindent
\textbf{\textit{2) Dilated gradient cues:}} Since pixel-wise gradients are sensitive to noise, we compute the regional energy $E_k(x,y)=\frac{1}{|\Omega|}\sum_{(u,v)\in\Omega(x,y)}\Gamma_k(u,v)$ to incorporate neighborhood statistics and enforce spatial consistency:

\begin{equation}
    C^{\text{C-LUT}}_{DG} = \mathcal{N}_{\mathcal{S}_C}\big(E_{near} - E_{far}\big).
\end{equation}

\noindent
\textbf{\textit{3) Maximum gradient cues:}} This cue evaluates the overall gradient magnitude across the two images, capturing local texture strength and reducing ambiguity in low-texture regions:
\begin{equation}
    C^{\text{C-LUT}}_{MG} = \mathcal{N}_{\mathcal{S}_C}\big(\max(\Gamma_{near}, \Gamma_{far})\big).
\end{equation}

\noindent
\textbf{\textit{4) Learnable regional semantic cues:}} We employ a lightweight CNN, denoted as $\mathcal{F}_{ctx}$, to extract joint contextual representations from the concatenated images, complementing gradient-based cues with region-level semantic information:
\begin{equation}
    C^{\text{C-LUT}}_{RS} = \mathcal{N}_{\mathcal{S}_C}\big(\mathcal{F}_{ctx}([I'_{near}, I'_{far}])\big).
\end{equation}

To optimize the efficiency of system integration, we implement a strategy that transforms large-scale computational tasks into offline look-up table operations. Building upon the establishment of the four cues, we utilize a four-dimensional LUT, denoted as $\mathcal{L_{C}}$, to map the continuous coordinate space $\mathbf{c} = \{c_1, c_2, c_3, c_4\} = \{C^{\text{C-LUT}}_{GD}, C^{\text{C-LUT}}_{DG},  C^{\text{C-LUT}}_{MG}, C^{\text{C-LUT}}_{RS}\}$ to the fusion weight:
\begin{equation}
\label{eq:lut_map}
\mathcal{M}_{main}^{LUT}(x, y) = \Phi_{\text{Main-LUT}} (\mathbf{c}),
\end{equation}
where $\Phi_{\text{Main-LUT}}$ represents the discrete look-up and interpolation operation. The LUT stores fusion weights in a four-dimensional lattice.

Since the calculated cues $\mathbf{c}$ are continuous floating-point values while the LUT grid is discrete, direct indexing is infeasible. Therefore, each coordinate $c_i$ ($i \in \{1, 2, 3, 4\}$) is decomposed into an integer grid index $n_i = \lfloor c_i \rfloor$ and a fractional residual $\alpha_i = c_i - n_i$, where $\lfloor \cdot \rfloor$ denotes the floor function.

To guarantee differentiability for gradient back-propagation and ensure smooth weight transitions, we employ Quadrilinear Interpolation. The output at an arbitrary query point is computed by aggregating the values from the $2^4=16$ surrounding vertices of the activated hypercube:
\begin{equation}
    \begin{aligned}
        \label{eq:interp_sum}
        \Phi_{\text{Main-LUT}}(\mathbf{c}) &= \sum_{\delta_1=0}^{1} \sum_{\delta_2=0}^{1} \sum_{\delta_3=0}^{1} \sum_{\delta_4=0}^{1} \mathcal{W}_{\boldsymbol{\delta}} \cdot \mathcal{L_{C}}(n'_1, n'_2, n'_3, n'_4),\\
        n'_a &= n_a + \delta_a, \quad a \in \{1, 2, 3, 4\},
    \end{aligned}
\end{equation}
where $\boldsymbol{\delta} = (\delta_1, \delta_2, \delta_3, \delta_4)$ represents the binary vertex offset. The interpolation weight $\mathcal{W}_{\boldsymbol{\delta}}$ is derived from the product of linear distances along each dimension:
\begin{equation}
\label{eq:interp_weight}
\mathcal{W}_{\boldsymbol{\delta}} = \prod_{j=1}^{4} \left( (1 - \alpha_j)^{1-\delta_j} \cdot \alpha_j^{\delta_j} \right).
\end{equation}

\noindent
\textbf{Detail-Edge Lookup Table (D-LUT):} The C-LUT captures semantic consistency and performs large-scale regional decisions. However, due to the limited resolution at the coarse scale, its capability to delineate fine-grained focus boundaries is inherently constrained. As a complement, the D-LUT is introduced to function as a high-resolution refiner. 
Crucially, while the C-LUT relies on first-order Sobel operators to capture broad structural trends and regional energy, the D-LUT utilizes the second-order Laplacian operator. Second-order responses emphasize rapid intensity transitions and zero-crossing structures, making them particularly suitable for boundary localization that first-order gradients often over-smooth. By employing the Laplacian convolution kernel $\mathcal{H}_{Lap}$, we directly extract these high-frequency structural responses $\Lambda_{near}$ and $\Lambda_{far}$:
\begin{equation}
    \Lambda_k = |\mathcal{H}_{Lap} \ast I_k|, \quad k \in \{near, far\}.
\end{equation}

Using a corresponding normalization operator $\mathcal{N}_{\mathcal{S}_D}(\cdot)$ adapted to the D-LUT bin size $\mathcal{S}_D$, we formulate the coordinate indices $\mathbf{d} = \{d_1, d_2\}$:

\vspace{2mm}
\noindent
\textbf{\textit{1) Laplacian difference cues:}} This cue computes the pixel-wise difference between second-order Laplacian responses of the two images, directly capturing high-frequency focus transitions and refining boundaries weakened during coarse-scale down-sampling:
\begin{equation}
    C^{\text{D-LUT}}_{LD} = \mathcal{N}_{\mathcal{S}_D}\big(\Lambda_{near} - \Lambda_{far}\big).
\end{equation}

\noindent
\textbf{\textit{2) Maximum Laplacian cues:}} This cue evaluates the maximum Laplacian magnitude across the two images, estimating local structural strength and suppressing unstable responses in flat or texture-scarce regions:

\begin{equation}
    C^{\text{D-LUT}}_{ML} = \mathcal{N}_{\mathcal{S}_D}\big(\Lambda_{\max}\big), \quad \Lambda_{\max} = \max(\Lambda_{near}, \Lambda_{far}).
\end{equation}

Building upon these two cues, a two-dimensional LUT, denoted as $\mathcal{L}_{D}$, maps the continuous coordinate space $\mathbf{d} = \{C^{\text{D-LUT}}_{LD}, C^{\text{D-LUT}}_{ML}\}$ to a high-frequency decision residual:
\begin{equation}
\label{eq:res_lut_map}
\Delta M = \Phi_{\text{Detail}}(\mathbf{d}).
\end{equation}

To maintain mathematical rigor, we define a boundary truncation operator $\mathcal{B}_{[a, b]}(z) = \max(a, \min(z, b))$ and decompose the continuous coordinates $d_k$ ($k \in \{1, 2\}$) into integer indices $m_k = \lfloor d_k \rfloor$ and fractional residuals $\beta_k = d_k - m_k$. Given $\delta_k \in \{0, 1\}$ denoting the binary vertex offsets. We determine the valid grid indices for the nearest neighbors as $m'_k = \mathcal{B}_{[0, \mathcal{S}_D-1]}(m_k + \delta_k)$, which explicitly confines the look-up operations within the LUT dimensions. To enable end-to-end training, we employ bilinear interpolation, aggregating the values of these $4$ nearest neighbors in the two-dimensional lattice:
\begin{equation}
    \begin{aligned}
        \Phi_{\text{Detail}}(\mathbf{d}) = \sum_{\delta_1=0}^{1} \sum_{\delta_2=0}^{1} \mathcal{W}_{\boldsymbol{\delta}} \cdot \mathcal{L}_{D}(m'_1, m'_2), \quad \mathcal{W}_{\boldsymbol{\delta}} = \prod_{j=1}^{2} \left( (1 - \beta_j)^{1-\delta_j} \cdot \beta_j^{\delta_j} \right).
    \end{aligned}
\end{equation}

\noindent
\textbf{Cross-Scale Map Integration.}
Through the explicit scale-decoupling of UMF-LUT, the final decision map $M_{final}$ is effortlessly obtained by integrating the semantically accurate coarse-region decision map from the C-LUT with the detail-edge decision map from D-LUT:
\begin{equation}
    M_{final} = \mathcal{B}_{[0, 1]}\big(\mathcal{D}_{\uparrow s}(\mathcal{M}_{main}^{LUT}) + \Delta M\big),
\end{equation}
where $\mathcal{D}_{\uparrow s}(\cdot)$ represents the bilinear up-sampling operator. The fused image $I_f$ is then generated via pixel-wise weighted aggregation:
\begin{equation}
    I_f = M_{final} \cdot I_{near} + (1 - M_{final}) \cdot I_{far}.
\end{equation}

\subsection{Loss Functions}
\label{subsec4}

To effectively optimize the scale-decoupled UMF-LUT architecture, we formulate specific learning objectives tailored to the distinct roles of the C-LUT and D-LUT. The respective loss functions are defined as:
\begin{equation}
    \begin{aligned}
        \mathcal{L}_{\text{C-LUT}} = \mathcal{L}_{rec} + \lambda_{1}\mathcal{L}_{bin}, \quad \mathcal{L}_{\text{D-LUT}} = \mathcal{L}_{rec} + \lambda_{2}\mathcal{L}_{sparse},
    \end{aligned}
\end{equation}
where $\mathcal{L}_{rec} = \lambda_{3}\mathcal{L}_{int} + \lambda_{4}\mathcal{L}_{grad} + \lambda_{5}\mathcal{L}_{ssim}$ denotes the fundamental unsupervised reconstruction loss, comprising standard intensity, gradient, and structural similarity constraints. $\lambda_{1} \sim \lambda_{5}$ are balancing hyperparameters.

To guide the network in achieving optimal scale-decoupling, we introduce two specialized regularization terms:

\vspace{2mm}
\noindent
\textbf{\textit{Binarization Loss.}} To prevent the C-LUT from outputting ambiguous intermediate values in flat regions and to suppress ghosting artifacts, this term forces the coarse decision map to make decisive selections:
\begin{equation}
    \mathcal{L}_{bin} = \mathbb{E} \big[ \mathcal{M}_{\text{coarse}} \odot (1 - \mathcal{M}_{\text{coarse}}) \big],
\end{equation}
where $\odot$ denotes element-wise multiplication, and $\mathbb{E}[\cdot]$ calculates the spatial expectation.

\vspace{2mm}
\noindent
\textbf{\textit{Sparsity Loss.}} To prevent the D-LUT from hallucinating pseudo-textures or amplifying noise, this regularization ensures that the high-frequency residual mask $\Delta M$ remains zero-dominant, activating exclusively at significant focal boundaries. We constrain this using the $\ell_1$-norm:
\begin{equation}
    \mathcal{L}_{sparse} = \|\Delta M\|_1.
\end{equation}

\section{UHD-MFF Dataset}
\label{sec:blind}
\begin{figure*}[t]
    \centering
    \begin{minipage}[t]{0.45\textwidth}
        \centering
        \captionof{table}{Comparison of our proposed UHD-MFF with existing multi-focus image fusion datasets. (Resolution corresponds to the average of the dataset, if not constant.)}
        \label{tab:dataset_comparison}
        \setlength{\tabcolsep}{1pt}
        \resizebox{1\linewidth}{!}{
        \begin{tabular}{c|cccc}
            \toprule[1.5pt]
            \rowcolor{black!10}
            \textbf{Dataset} & \textbf{Source} & \textbf{Resolution} & \textbf{Pairs} &  \textbf{GT} \\
            \midrule
            Classic~\cite{liu2020multi-survey}  & Real      & $512 \times 512$   & 10    & \xmark \\
            \rowcolor{black!10}
            Lytro~\cite{nejati2015multi}    & Real      & $520 \times 520$   & 20    & \xmark \\
            MFFW2~\cite{xu2020mffw}    & Synthetic & $600 \times 400$   & 13    & \cmark \\
            \rowcolor{black!10}
            MFI-WHU~\cite{zhang2021mff}  & Synthetic  & $600 \times 400$   & 120   & \cmark \\
            Real-MFF~\cite{zhang2020real} & Real    & $625 \times 433$   & 710   & \xmark \\
            \midrule
            \rowcolor{black!10}
            \textbf{UHD-MFF-Real}& \textbf{Real}     & $\mathbf{3840 \times 2160}$ & \textbf{150}   & \xmark \\
            \rowcolor{white!50}
            \textbf{UHD-MFF-Syn} & \textbf{Synthetic} & $\mathbf{3840 \times 2160}$ & \textbf{1,800} & \cmark \\
            \bottomrule[1.5pt]
        \end{tabular}
        }
    \end{minipage}
    \hfill
    \begin{minipage}[t]{0.48\textwidth}
        \centering
        \vspace{0.6em}
        \includegraphics[width=\linewidth]{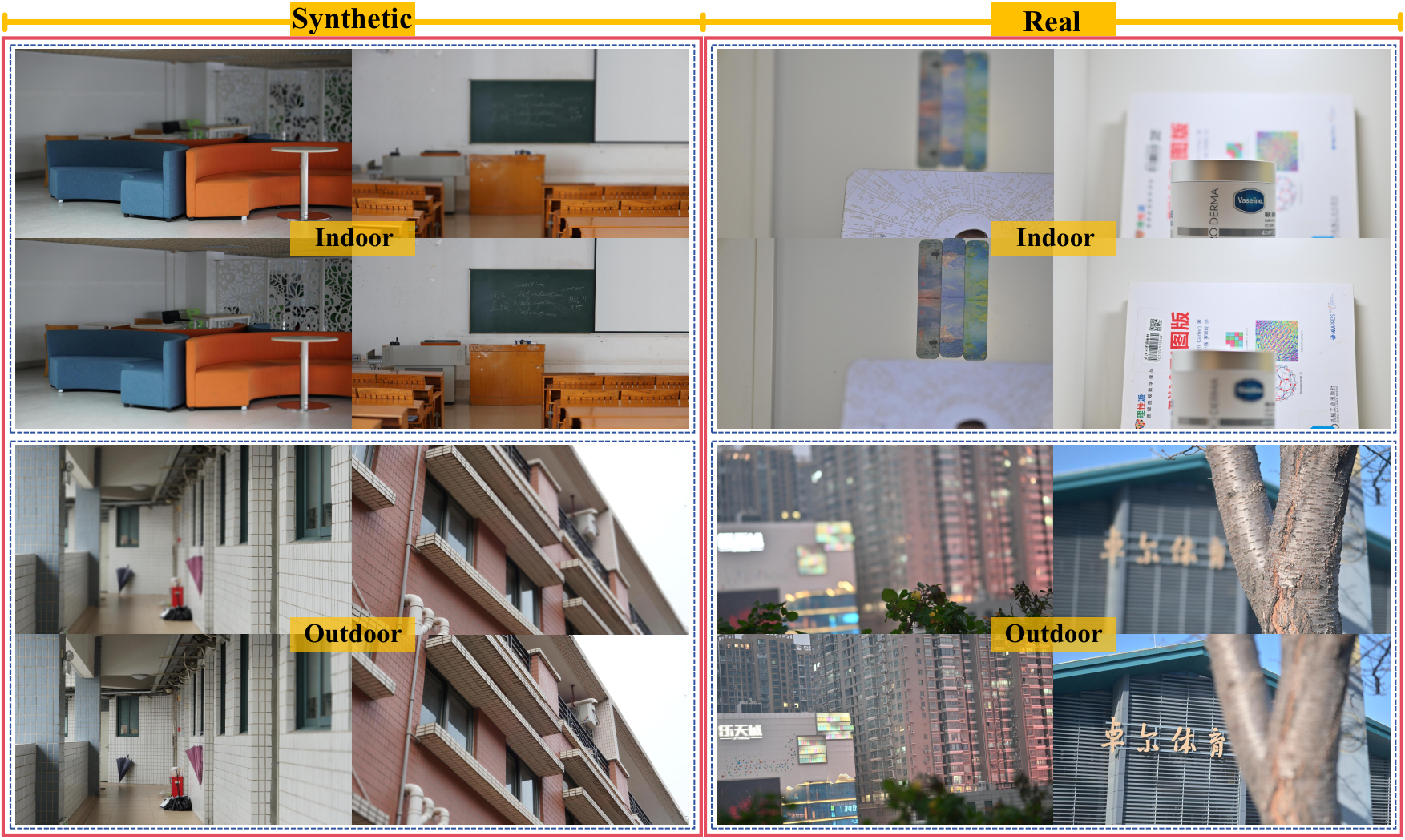}
        \vspace{-0.2in}
        \captionof{figure}{Samples from UHD-MFF.}
        \label{fig:4-1}
    \end{minipage}
\end{figure*}

To overcome the data availability barrier, we establish the first ultra-high-definition multi-focus fusion dataset, termed as UHD-MFF, which contains 1,950 pairs of UHD images with diverse and complex scenes. As shown in Tab.~\ref{tab:dataset_comparison}, UHD-MFF dataset demonstrates strong competitiveness compared with existing mainstream datasets in terms of both resolution and scale. Representative scenes are illustrated in Fig.~\ref{fig:4-1}. It consists of UHD-MFF-Real and UHD-MFF-Syn subsets.

\vspace{2mm}
\noindent
\textbf{UHD-MFF-Real.} To validate generalization in practical scenarios, we meticulously captured 150 real-world image pairs at $3840 \times 2160$ resolution. Strict pixel-level alignment was ensured during acquisition. Comprehensive capture protocols are deferred to the Supplementary Material.

\begin{figure*}[t!]
\vspace{0.1in}
\centering
\includegraphics[width=0.98\linewidth]{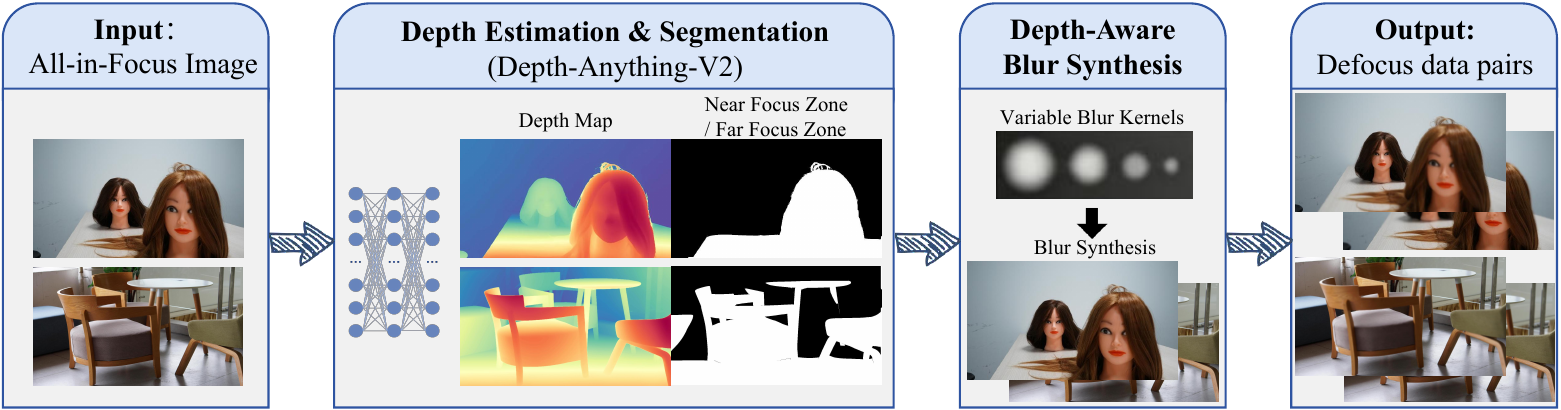}
\caption{UHD-MFF-Syn dataset synthesis pipeline.}
\label{fig:4-2}
\end{figure*}

\vspace{2mm}
\noindent
\textbf{UHD-MFF-Syn.} Utilizing high-quality all-in-focus images from the UHD-LL dataset~\cite{Li2023uhdfour}, we adopted a depth-guided variable blur strategy to simulate realistic defocus effects. The simulation process is illustrated in Fig.~\ref{fig:4-2}. After quality filtering, the 1,800 synthetic pairs were randomly divided into 1,500 for training and 300 for testing.

\section{Experiment}

\subsection{Experiment Settings}
\label{subsec1}
\noindent
\textbf{Implementation Details.} C-LUT and D-LUT are jointly optimized within a unified training framework using the AdamW optimizer. The batch size is set to 8, and the initial learning rate is fixed at $1\times10^{-5}$. The loss function is formulated as a weighted combination of multiple terms, with coefficients $\lambda_{1}=50$, $\lambda_{2}=0.05$, $\lambda_{3}=20$, $\lambda_{4}=0.7$, and $\lambda_{5}=0.8$. Experiments are conducted in PyTorch on an NVIDIA RTX 3090, with mobile deployment tested on a Huawei P60 Pro.

\vspace{2mm}
\noindent
\textbf{Datasets.} We evaluate our method on our proposed UHD-MFF dataset. To assess zero-shot generalization, the model trained exclusively on UHD-MFF-Syn is directly evaluated on UHD-MFF-Real dataset.

\vspace{2mm}
\noindent
\textbf{Metrics.} We evaluate the performance of the image fusion models using several widely adopted objective metrics, including standard deviation (SD), sum of correlation differences (SCD), mutual information (MI)~\cite{qu2002information}, information entropy (EN)~\cite{roberts2008assessment}, visual information fidelity (VIF), quality of gradient-based fusion ($Q^{AB/F}$)~\cite{ma2019qabf}, average gradient (AG), and spatial frequency (SF). Higher values of all these metrics indicate higher quality of the fusion image.

\vspace{2mm}
\noindent
\textbf{SOTA Competitors.} To validate the effectiveness of the proposed approach, we performed comprehensive comparisons with multiple state-of-the-art image fusion methods on multiple datasets. The compared methods include task specific fusion approaches, e.g. SMFuse~\cite{ma2021smfuse}, SESF~\cite{ma2021sesf}, DRPL~\cite{li2020drpl}, MSFIN-Fusion~\cite{liu2021multiscale}, Fusion2Void~\cite{lin2024fusion2void}, CCSR-Net~\cite{zheng2025unfolding}, FusionGCN~\cite{ouyang2025fusiongcn} and unified image fusion frameworks, e.g. GIFNet~\cite{Cheng_2025_CVPR}, TC-MoA~\cite{Zhu_2024_CVPR}.

\subsection{Qualitative Experiments}
\label{subsec2}

\begin{figure*}[t!]
\centering
\includegraphics[width=0.99\linewidth]{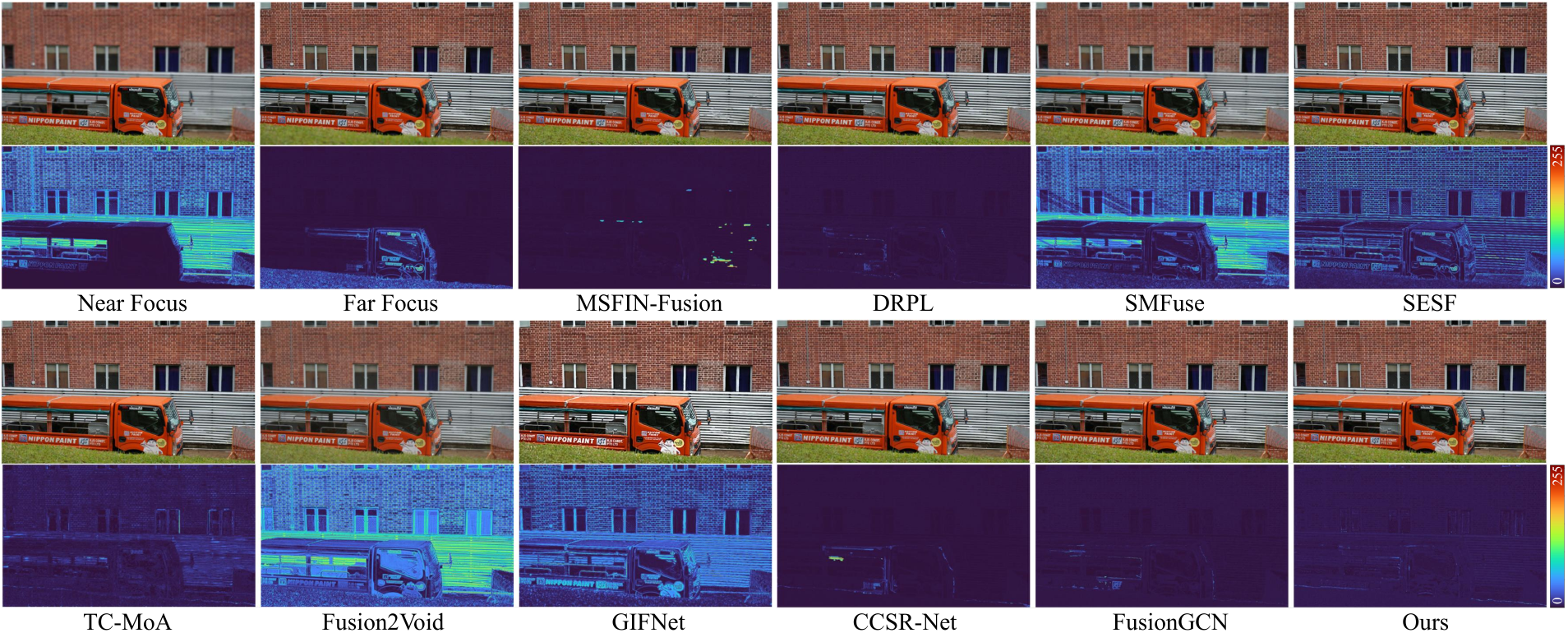}
\caption{Qualitative comparison on UHD-MFF-Syn.}
\label{fig:5-2-1}
\end{figure*}

\begin{figure*}[t!]
\vspace{0.1in}
\centering
\includegraphics[width=0.99\linewidth]{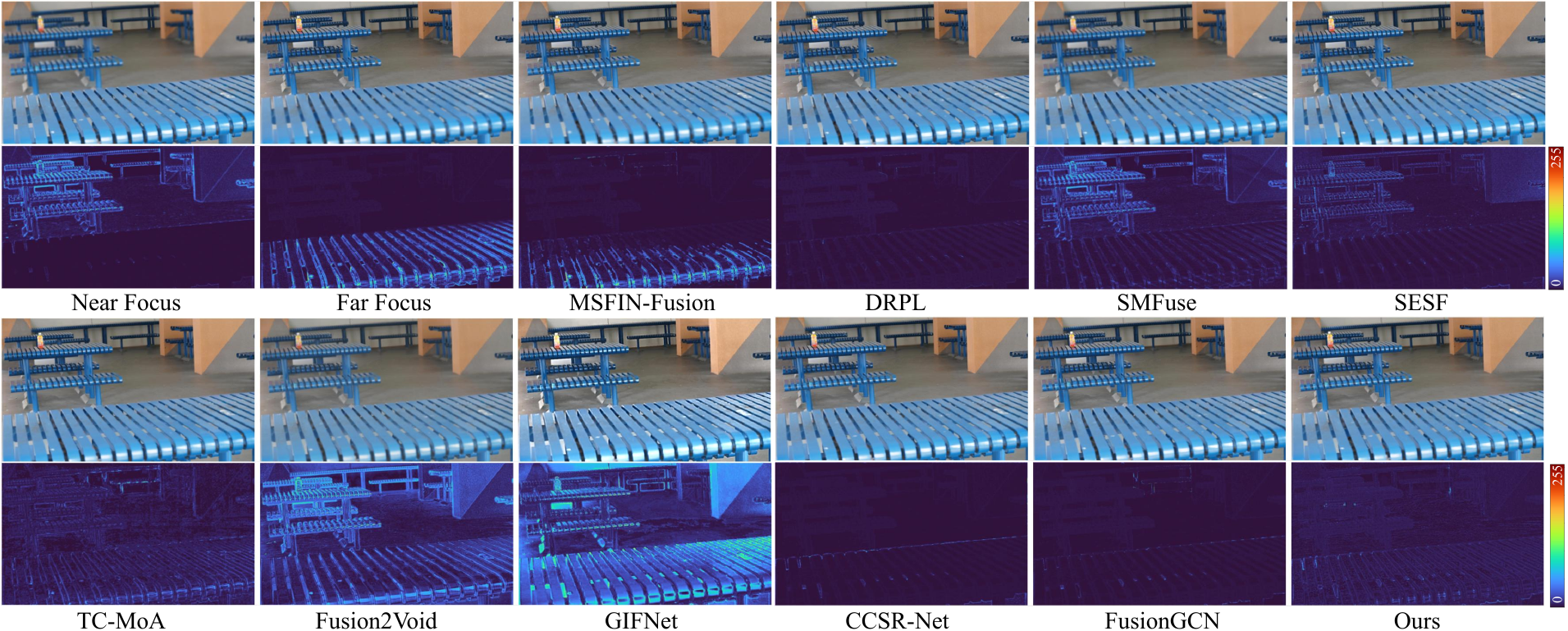}
\caption{Qualitative comparison on UHD-MFF-Syn.}
\label{fig:5-2-2}
\end{figure*}

\begin{figure*}[t!]
\centering
\includegraphics[width=0.99\linewidth]{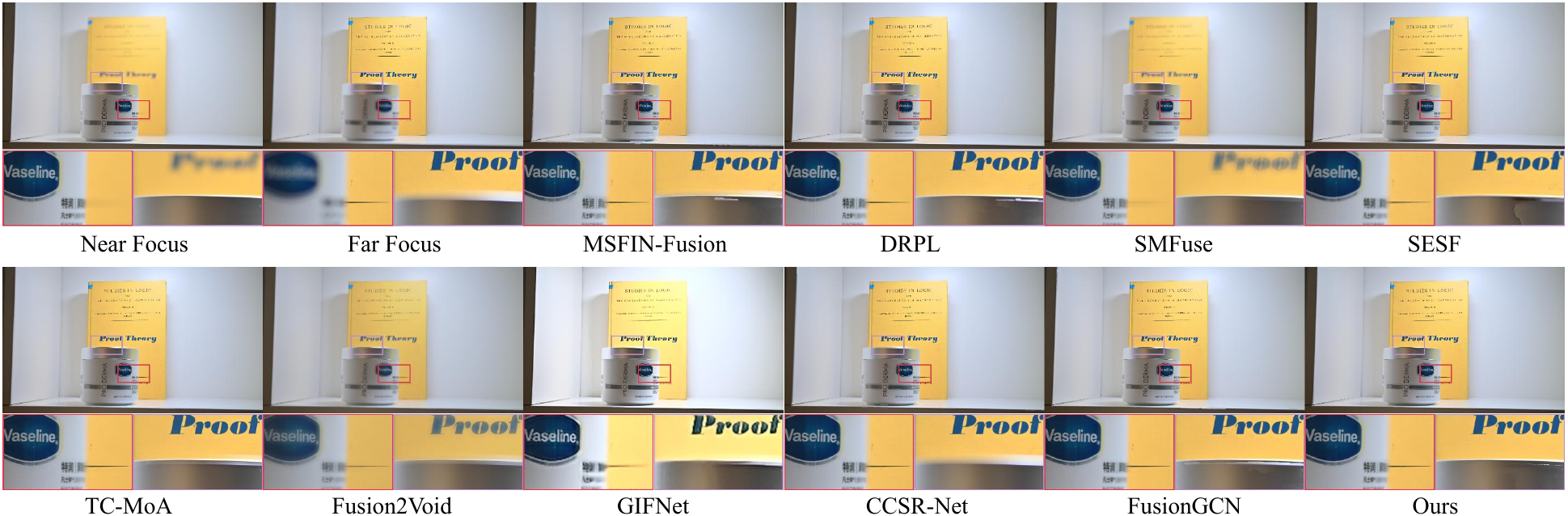}
\caption{Qualitative comparison on UHD-MFF-Real.}
\label{fig:5-2-4}
\end{figure*}

We qualitatively compare our method with nine state-of-the-art approaches across UHD-MFF datasets. The visual results are illustrated in Figs.~\ref{fig:5-2-1}-~\ref{fig:5-2-4}. Among them, Fig.~\ref{fig:5-2-1} and Fig.~\ref{fig:5-2-2} display the results on the UHD-MFF-Syn dataset. Observing the residual maps, MSFIN-Fusion struggles to preserve sufficient near-focus information, whereas SMFuse and SESF suffer from severe detail loss in far-focus regions. TC-MoA exhibits a relatively uniform information loss across the entire image.
Fig.~\ref{fig:5-2-4} visualizes the results on the UHD-MFF-Real dataset. Although GIFNet retains details reasonably well, its fused results suffer from significant luminance and color distortions. As clearly seen in Fig.~\ref{fig:5-2-4}, GIFNet introduces noticeable brightness shifts and color deviations during the fusion process. At the foreground-background boundary around the bottle cap, our method demonstrates the best detail preservation capability, while other methods introduce varying degrees of artifacts or blurring. Furthermore, our approach achieves the best overall texture preservation across the entire image. Comprehensive evaluations across the two datasets indicate that our method exhibits distinct advantages on high-resolution real-world images while maintaining robust performance on low-resolution inputs.

\begin{table*}[t]
\vspace{0.1in}
    \caption{Quantitative comparisons on the UHD-MFF dataset. \textbf{Bold} indicates the best, and \underline{underline} indicates the second-best.}
    \label{tab:syn-real-combined}
    \centering
    \setlength{\tabcolsep}{2pt}
    \renewcommand{\arraystretch}{1.1}
    \resizebox{1\textwidth}{!}{
    \begin{tabular}{l| cccccc c| cccccc c}
        \toprule
        \multirow{2}{*}{\textbf{Method}} &
        \multicolumn{7}{c|}{\textbf{UHD-MFF-Real}} &
        \multicolumn{7}{c}{\textbf{UHD-MFF-Syn}} \\
        \cmidrule(lr){2-8} \cmidrule(lr){9-15}
        & \textbf{SD}↑ & \textbf{MI}↑ & \textbf{EN}↑ & \textbf{VIF}↑ & $\mathbf{Q^{AB/F}}$↑ & \textbf{AG}↑ & \textbf{SF}↑ & \textbf{SD}↑ & \textbf{MI}↑ & \textbf{EN}↑ & \textbf{VIF}↑ & $\mathbf{Q^{AB/F}}$↑ & \textbf{AG}↑ & \textbf{SF}↑ \\
        \midrule
        MSFIN-Fusion & 41.0410 & \underline{4.7667} & \underline{6.9540} & 1.3038  & \underline{0.7145} & 2.9993 & 8.6925 & 49.5343 & \textbf{6.7902} & 7.1517 & 1.3781 & 0.6280 & 1.4852 & 5.1240 \\
        DRPL & 41.2584 & 4.5540 & 6.8674 & 1.3187 & 0.6938 & 2.9273 & 8.8130 & 49.0376 & 6.1102 & 7.1611 & 1.4034 & 0.7121 & \underline{1.9383} & \underline{6.7343} \\
        SMFuse & 39.2386 & 3.5286 & 6.8001 & 0.8858 & 0.3739 & 1.6319 & 4.7578 & 48.8134 & 5.6906 & 7.1196 & 1.1449 & 0.5046 & 1.1714 & 4.0319 \\
        SESF  & 41.0118 & 4.6523 & 6.7696 & \underline{1.3628} & 0.7015 & \underline{3.0181} & 8.7894 & 49.5925 & 5.3177 & 7.1522 & 0.9414 & 0.3545 & 1.1705 & 3.3623 \\
        TC-MoA & 40.8527 & 3.8540 & 6.7344 & 1.2048 & 0.6828 & 2.9395 & 8.6218 & 49.8702 & 5.9282 & \underline{7.1661} & 1.4106 & \underline{0.7491} & 1.9318 & 6.7099 \\
        Fusion2Void & 34.3669 & 3.4135 & 6.6375 & 0.8356 & 0.3535 & 1.6709 & 5.1135 & 42.9653 & 5.2050 & 6.9463 & 0.9647 & 0.4350 & 1.2541 & 4.5707 \\
        GIFNet & \textbf{43.6695} & 3.0130 & 6.7784 & 0.4957 & 0.1952 & 2.0124 & 5.4410 & 45.5961 & 4.4482 & 7.0551 & 0.8195 & 0.2674 & 1.5112 & 5.1649 \\
        \rowcolor{blue!10}
        CCSR-Net & 41.6663 & 4.5768 & 6.9491 & 1.3410 & 0.7002 & 2.9764 & \underline{8.7278} & \underline{49.8757} & 6.2169 & 7.1659 & \underline{1.4186} & 0.7448 & 1.9293 & 6.6790 \\
        FusionGCN & 40.6867 & 4.6535 & 6.9472 & 1.3147 & 0.7078 & 2.8628 & 8.6057 & 49.4332 & \underline{6.7128} & 7.1621 & 1.4036 & 0.7216 & 1.9163 & 6.6971 \\
        \rowcolor{pink!50}
        \textbf{Ours}  & \underline{41.8366} & \textbf{4.8029} & \textbf{6.9588} & \textbf{1.3814} & \textbf{0.7290} & \textbf{3.0344} & \textbf{8.8598} & \textbf{49.9062} & 6.3999 & \textbf{7.1672} & \textbf{1.4335} & \textbf{0.7498} & \textbf{1.9416} & \textbf{6.7443} \\
        \bottomrule
    \end{tabular}
    }
\end{table*}

\subsection{Quantitative Experiments}
\label{subsec3}
The quantitative comparisons between our method and other SOTA approaches across the UHD-MFF dataset are reported in Tab.~\ref{tab:syn-real-combined}. As shown, our method achieves the best performance in six out of seven evaluated metrics. On the UHD-MFF-Syn, in terms of SD and VIF, our approach outperforms the second-best method, CCSR-Net, by margins of 0.0305 and 0.0149, respectively. For EN and $Q^{AB/F}$, our method surpasses the runner-up, TC-MoA, by 0.0011 and 0.0007. These quantitative results verify that our fusion strategy retains significantly more source image information compared to other competing methods. On the UHD-MFF-Real, our method exhibits a comprehensive advantage over the alternatives, further demonstrating our outstanding capability in handling high-resolution fusion tasks.

\subsection{Ablation Experiments}
\label{subsec4}
We conduct comprehensive ablation studies on the UHD-MFF benchmark to validate our architectural components, training data scale, and supervision paradigms.

\begin{figure*}[t!]
\centering
\includegraphics[width=0.85\linewidth]{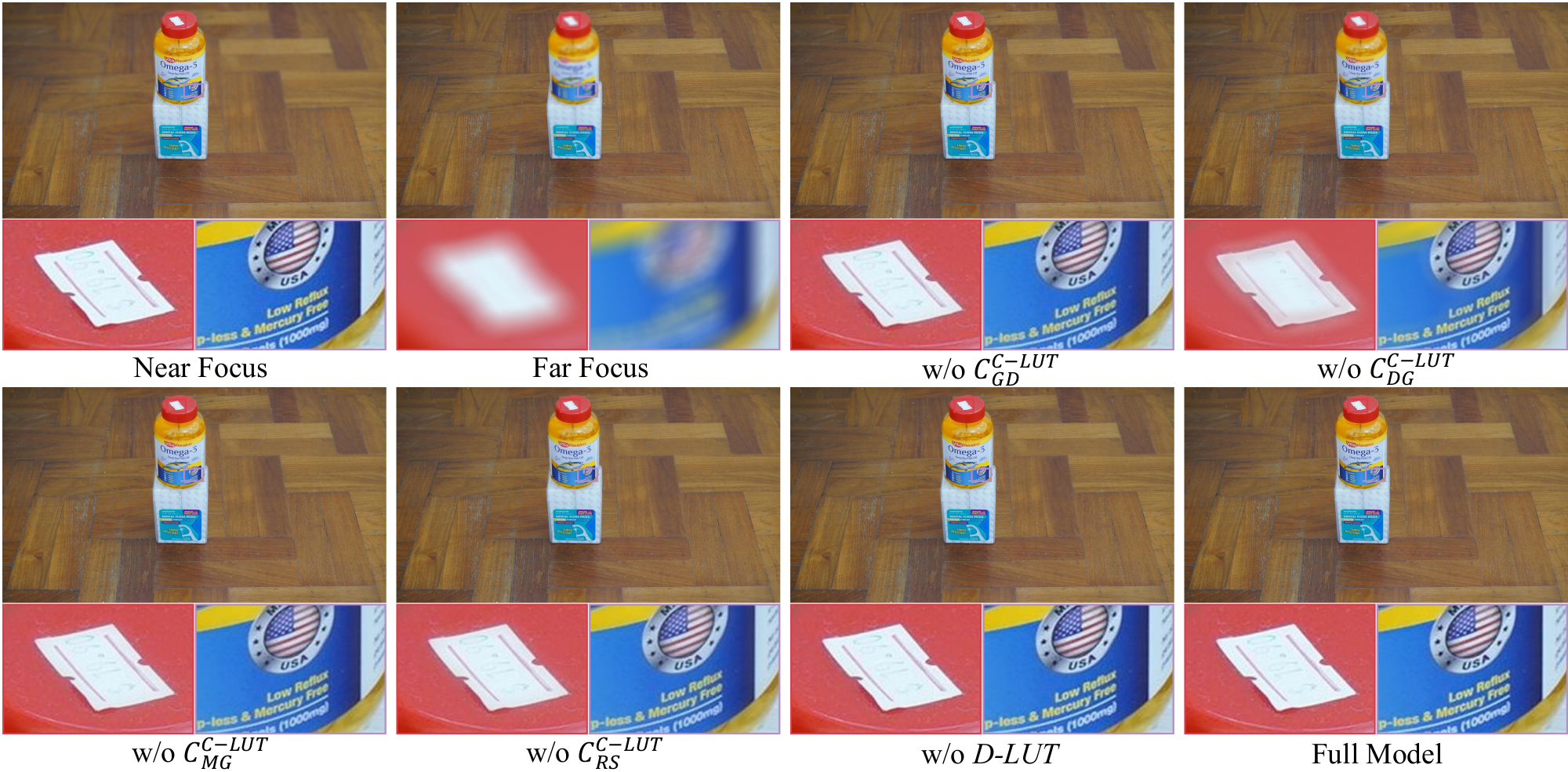}
\caption{Qualitative comparison of ablation experiment on UHD-MFF dataset.}
\label{fig:abla-qual}
\end{figure*}

\vspace{2mm}
\noindent
\textbf{Effectiveness of Individual Components.} We evaluate the four focus cues and the Detail Branch. As shown in Tab.~\ref{tab:ablation-quan} and Fig.~\ref{fig:abla-qual}, removing any single module degrades fusion quality, causing boundary artifacts and detail loss. Our full model consistently achieves the best quantitative performance, confirming the indispensability of each component.

\begin{table*}[t!]
\vspace{0.1in}
  \centering
  \caption{Ablation study of different components.}
  \label{tab:ablation-quan}
  \setlength{\tabcolsep}{8pt}
  \resizebox{0.98\textwidth}{!}{
    \begin{tabular}{c|cccccccc}
      \toprule
      \textbf{Model Type} & \textbf{SD}↑ & \textbf{SCD}↑ & \textbf{MI}↑ & \textbf{EN}↑ & \textbf{VIF}↑ & $\mathbf{Q^{AB/F}}$↑ & \textbf{AG}↑ & \textbf{SF}↑ \\
      \midrule
      w/o \small$C^{\text{C-LUT}}_{GD}$ & 49.7349 & 0.3527 & 6.1312 & 7.1586 & 1.4028 & 0.7449 & 1.7748 & 6.1405 \\
      w/o \small$C^{\text{C-LUT}}_{DG}$ & 49.2397 & 0.2412 & 5.9196 & 7.1497 & 1.2564 & 0.6601 & 1.5416 & 5.4037 \\
      w/o \small$C^{\text{C-LUT}}_{MG}$ & 49.6651 & 0.3407 & 6.0500 & 7.1580 & 1.3948 & 0.7436 & 1.7491 & 6.0278 \\
      w/o \small$C^{\text{C-LUT}}_{RS}$ & 49.5990 & 0.3351 & 5.9962 & 7.1564 & 1.3833 & 0.7417 & 1.7202 & 5.8832 \\

      w/o \small{D-LUT} & 49.8771 & \textbf{0.3540} & 6.3538 & 7.1636 & 1.4210 & \textbf{0.7501} & 1.8479 & 6.4938\\
      \midrule
      \rowcolor{pink!50}
      \textbf{Full Model} & \textbf{49.9062} & 0.3414 & \textbf{6.3999} & \textbf{7.1672} & \textbf{1.4335} & 0.7498 & \textbf{1.9416} & \textbf{6.7443} \\
      \bottomrule
    \end{tabular}
    }
\end{table*}

\begin{figure*}[t!]
\centering
\includegraphics[width=1\linewidth]{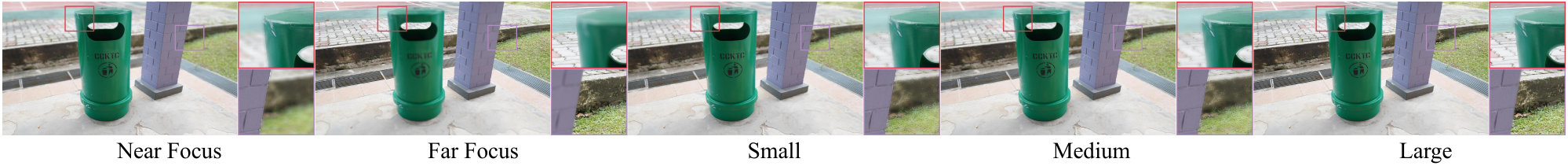}
\caption{Qualitative comparison of different training set sizes.}
\label{fig:abla-datascal}
\end{figure*}

\vspace{2mm}
\noindent
\textbf{Impact of Training Data Scale.}
To evaluate the impact of dataset scale, we conduct an ablation study with three training sizes: 1500 (large), 150 (medium), and 15 (small). As shown in Fig.~\ref{fig:abla-datascal} and Tab.~\ref{tab:training_size}, performance exhibits a consistent upward trend with increasing data scale. All quantitative metrics improve progressively as the training set expands, validating large-scale necessity.

\begin{table}[t!]
  \centering
  \caption{Quantitative comparison of different training set sizes.}
  \label{tab:training_size}
  \setlength{\tabcolsep}{7pt}
  \resizebox{0.94\textwidth}{!}{
  \begin{tabular}{c|ccccccccc}
      \toprule
      \textbf{Scale of Data}  & \textbf{AG}↑ & \textbf{SD}↑ & \textbf{SCD}↑ & \textbf{MI}↑ & \textbf{EN}↑ & \textbf{SF}↑ & \textbf{VIF}↑ & $\mathbf{Q^{AB/F}}$↑ \\
      \midrule
      Small     & 1.4792 & 49.0784 & 0.2223 & 6.1959 & 7.1415 & 6.4114 & 1.1637 & 0.5367 \\
      Medium   & 1.4904 & 49.1030 & 0.2329 & 6.1716 & 7.1423 & 6.3984 & 1.1726 & 0.5400 \\
      \rowcolor{pink!50}
      Large & \textbf{1.9416} & \textbf{49.9062} & \textbf{0.3414} & \textbf{6.3999} & \textbf{7.1672} & \textbf{6.7443} & \textbf{1.4335} & \textbf{0.7498} \\
      \bottomrule
  \end{tabular}
  }
\end{table}

\begin{figure*}[t!]
\vspace{0.1in}
\centering
\includegraphics[width=0.99\linewidth]{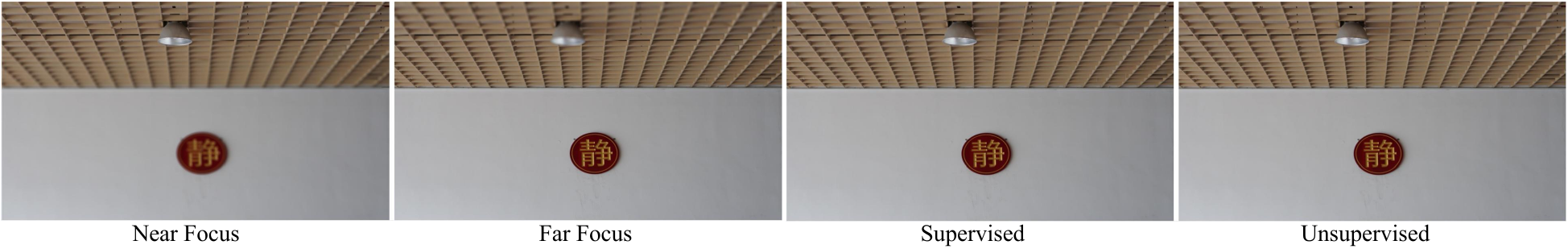}
\caption{Qualitative comparison between supervised and unsupervised strategies.}
\label{fig:abla-super}
\end{figure*}

\vspace{2mm}
\noindent
\textbf{Unsupervised vs. Supervised Paradigms.} Despite having synthetic ground-truth images, we deliberately employ an unsupervised strategy to prevent the network from overfitting to specific synthetic degradation patterns. Comparing our model with a fully supervised counterpart shown in Fig.~\ref{fig:abla-super} and Tab.~\ref{tab:sup_vs_unsup}, we observed that while numerical metrics are comparable, the unsupervised paradigm yields significantly more natural luminance and softer transitions. By avoiding the unnatural contrast artifacts occasionally seen in supervised results, our focus-aware unsupervised loss better preserves spectral consistency and generic sharpness discrimination, justifying it as our default setting.

\begin{table}[t!]
\vspace{0.1in}
  \centering
  \caption{Quantitative comparison between supervised and unsupervised strategies.}
  \label{tab:sup_vs_unsup}
  \setlength{\tabcolsep}{7pt}
  \resizebox{0.96\textwidth}{!}{
  \begin{tabular}{c|cccccccc}
      \toprule
      \textbf{Training Strategy} & \textbf{AG}↑ & \textbf{SD}↑ & \textbf{SCD}↑ & \textbf{MI}↑ & \textbf{EN}↑ & \textbf{SF}↑ & \textbf{VIF}↑ & $\mathbf{Q^{AB/F}}$↑ \\
      \midrule
      Supervised   & 1.9301 & 49.9301 & 0.3535 & 6.3918 & 7.1666 & 6.7423 & 1.4367 & 0.7491 \\
      Unsupervised & 1.9416 & 49.9062 & 0.3414 & 6.3999 & 7.1672 & 6.7443 & 1.4335 & 0.7498 \\
      \bottomrule
  \end{tabular}%
  }
\end{table}

\begin{figure*}[!t]
\vspace{0.1in}
\centering
\includegraphics[width=0.99\linewidth]{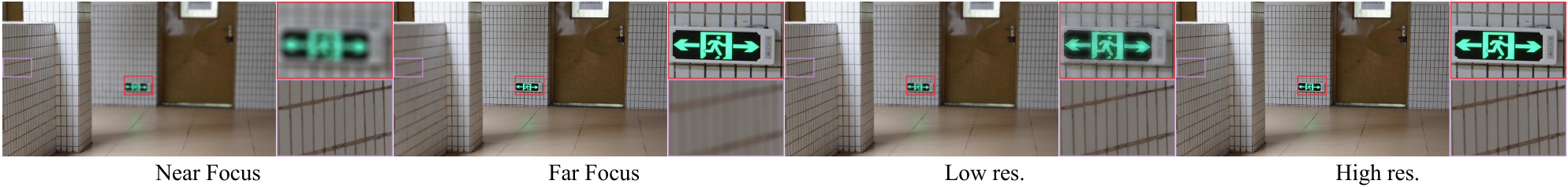}
\caption{Qualitative comparison of the model performance trained on low-resolution and high-resolution data.}
\label{fig:abla-resolution}
\end{figure*}

\vspace{2mm}
\noindent
\textbf{Impact of Training Resolution.} To verify the necessity of high-resolution training data, we trained our model on existing low-resolution datasets and evaluated its performance on UHD scenarios. As shown in Fig.~\ref{fig:abla-resolution} and Tab.~\ref{tab:resolution_generalization}, models trained on low-resolution data suffer from severe detail degradation when applied to 4K inputs. Both visual quality and quantitative metrics confirm that training on native 4K datasets is indispensable for high-fidelity UHD fusion.

\begin{table*}[!t]
\vspace{0.1in}
  \centering
  \caption{Performance comparison on the high resolution test set.}
  \label{tab:resolution_generalization}
  \setlength{\tabcolsep}{7pt}
  \resizebox{0.96\textwidth}{!}{
  \begin{tabular}{c| cccccccc}
    \toprule
    \textbf{Type of Data} & \textbf{AG}↑ & \textbf{SD}↑ & \textbf{SCD}↑ & \textbf{MI}↑ & \textbf{EN}↑ & \textbf{SF}↑ & \textbf{VIF}↑ & $\mathbf{Q^{AB/F}}$↑ \\
    \midrule
    low resolution  & 1.6032 & 49.2849 & 0.2886 & 5.8601 & 7.1475 & 6.2509 & 1.2233 & 0.6129 \\
    \rowcolor{pink!50}
    high resolution & \textbf{1.9416} & \textbf{49.9062} & \textbf{0.3414} & \textbf{6.3999} & \textbf{7.1672} & \textbf{6.7443} & \textbf{1.4335} & \textbf{0.7498} \\
    \bottomrule
  \end{tabular}
  }
\end{table*}

\begin{table*}[!t]
\vspace{0.1in}
    \centering
    \caption{Comparison of computational efficiency and model complexity.}
    \label{tab:efficiency_comparison}
    \setlength{\tabcolsep}{6pt}
    \resizebox{0.93\textwidth}{!}{
    \begin{tabular}{l|ccccc}
        \toprule
        \textbf{Method} & \makecell{\textbf{Test Time}\\ \textcolor{gray!150}{(ms)}} & \makecell{\textbf{FLOPs} \\ \textcolor{gray!150}{(G)}} & \makecell{\textbf{Params} \\ \textcolor{gray!150}{(M)}} & \makecell{\textbf{Peak Mem} \\ \textcolor{gray!150}{(MB)}} & \makecell{\textbf{Energy Cost} \\ \textcolor{gray!150}{(mJ)}}\\
        \midrule
        TC-MoA & $5714.91 \pm 13.78$ & 94370.32 & 340.89 & 12024.95 & 434155.48\\
        SMFuse & $\underline{150.35 \pm 20.96}$ & 637.57 & 0.038 & 29651.47 & 1514.18\\
        \rowcolor{blue!10}
        SESF & $450.85 \pm 10.06$ & \textbf{14.814} & \underline{0.014} & \underline{2109.36} & \underline{84.07}\\
        Fusion2Void & $883.48 \pm 16.49$ & 2161.85 & 0.26 & 7374.96 & 9992.3\\
        MSFIN-Fusion & $29453.10 \pm 4891.83$ & 1693.34 & 4.59 & 10573.32 & 7837.12\\
        GIFNet & $235.27 \pm 1.44$ & 381.56 & 0.82 & 8015.64 & 1771.11\\
        DRPL & $484.11 \pm 21.40$ & 8890.54 & 1.071 & 17744.65 & 40944.24\\
        CCSR-Net & $2910.29 \pm 409.93$ & \underline{50.21} & 0.025 & 6392.68 & 278.72\\
        FusionGCN & $15382.78 \pm 500.96 $ & 388.56 & 0.13 & 17350.42 & 1835.17 \\
        \midrule
        \rowcolor{pink!50}
        \textbf{Ours} & $\mathbf{11.03 \pm 0.82}$ & $ 67.82 $ & $\mathbf{0.0081}$ & $\mathbf{1133.15}$ & $\mathbf{67.45}$\\
        \bottomrule
    \end{tabular}
    }
\end{table*}

\vspace{2mm}
\noindent
\textbf{Performance Comparison.}
Superior computational efficiency constitutes the most distinct advantage of our proposed framework. The comparative performance benchmarks are detailed in Tab.~\ref{tab:efficiency_comparison}. Tested on an NVIDIA GeForce RTX 3090 GPU with the UHD-MFF dataset, our method achieves an average inference latency of 11.03 ms per image, which translates to a throughput of 90 frames per second. Notably, our model architecture is the most compact among all competing methods, containing merely 0.0081 million parameters. Regarding resource overhead during inference, our approach requires a peak memory of only 1133.15 MB and consumes a minimal energy of 67.45 mJ.

To validate the practicality of our method in real-world scenarios, we deployed the proposed framework onto a mobile platform. We selected the Huawei P60 Pro as our experimental testbed, which is equipped with the Snapdragon 8+ Gen 1 processor.
The performance comparison is presented in Tab.~\ref{tab:mobile_performance}.
Among the selected comparison methods, only SESF can run on a mobile device, with an average inference latency of 55.68 s, while the other methods fail to operate on mobile devices. In contrast, the average inference latency of our method consistently remains within 1 second, even for high-resolution inputs, demonstrating its substantial advantage for practical deployment on mobile devices.

\begin{table*}[!t]
  \centering
  \caption{Comparison of performance on mobile devices.}
  \label{tab:mobile_performance}
  \setlength{\tabcolsep}{6pt}
  \resizebox{0.82\textwidth}{!}{
    \begin{tabular}{l|ccccc}
      \toprule
      \textbf{Metrics} & TC-MoA & SMFuse & SESF & Fusion2Void & MSFIN-Fusion \\
      \midrule
      \textbf{Test time (s)} & $-$ & $-$ & 55.68 & $-$ & $-$ \\
      \textbf{Deployment}    & \xmark & \xmark & \cmark & \xmark & \xmark \\
      \midrule
      \textbf{Metrics} & GIFNet & DRPL & CCSR-Net & FusionGCN & \textbf{Ours} \\
      \midrule
      \textbf{Test time (s)} & $-$ & $-$ & $-$ & $-$ & \textbf{0.97} \\
      \textbf{Deployment}    & \xmark & \xmark & \xmark & \xmark & \cmark \\
      \bottomrule
    \end{tabular}
  }
\end{table*}

\section{Conclusion}
This paper effectively shatters the three major barriers, including data availability, model adaptability, and deployment feasibility, that hinder multi-focus image fusion in ultra-high-definition scenarios. To bridge the existing data gap, we establish UHD-MFF, the first large-scale UHD resolution multi-focus dataset, providing a robust foundation for future research. Furthermore, we propose UMF-LUT, a highly efficient, scale-specialized lookup-table framework tailored for UHD images. By innovatively decoupling the fusion process into low-resolution region-level decisions via C-LUT and high-resolution edge-level refinements via D-LUT, our method ensures accurate focus perception while successfully bypassing the massive memory overhead of full-resolution convolutions. Extensive experiments confirm that UMF-LUT not only achieves state-of-the-art performance in both visual fidelity and quantitative metrics, but also realizes real-time 4K fusion with minimal computational cost. Ultimately, this work breaks the hardware limitations of existing methods, demonstrating strong feasibility for smartphone deployment and significantly advancing the practical application of multi-focus image fusion in next-generation UHD imaging.

\section*{Acknowledgements}
This work was supported by the National Natural Science Foundation of China under Grant nos. 625B2135, and 62276192.

\bibliographystyle{splncs04}
\bibliography{main}
\end{document}